\documentclass[letterpaper, 10 pt, conference]{ieeeconf}  

\IEEEoverridecommandlockouts                              

\def \sysname {LookUP}
\def \bestFrameRate {5}
\def \bestAccuracy {1.2}
\def \GTAccuracy {5}
\def \datasetFrameNumber {276,063}
\def \datasetTimeSpan {5,117}
\def \datasetLength {50}
\def \coarseLocalisationUnitName {Semi-Supervised SLAM}

\title{\LARGE \bf
	\sysname{}: Vision-Only Real-Time Precise Underground Localisation for Autonomous Mining Vehicles
}

\author{Fan Zeng$^1$, Adam Jacobson$^1$, David Smith$^2$, Nigel Boswell$^2$, Thierry Peynot$^1$, Michael Milford$^1$
	\thanks{This research was supported by an Advance Queensland Innovation Partnerships grant from the Queensland Government, Mining3, Caterpillar and the Queensland University of Technology (QUT). MM also received support from an ARC Future Fellowship FT140101229.
	}
	\thanks{$^1$FZ, AJ, TP and MM are with QUT
		{\tt\small fan.zeng@qut.edu.au}}
	\thanks{$^2$DS and NB are with Caterpillar, Inc.}
}
\usepackage{amsmath}
\usepackage{amssymb}
\usepackage[noend]{algorithm2e}

\usepackage{graphicx}
\usepackage{booktabs}
\usepackage{subcaption}
\IEEEoverridecommandlockouts

\usepackage{tikz}
\usepackage{textcomp}
\usepackage{hyperref}
\usepackage{lipsum}
\newcommand\copyrighttext{%
	\footnotesize \textcopyright This  paper  is  a  preprint  (IEEE  “accepted”  status). IEEE copyright notice. \textcopyright 2019 IEEE.  Personal use of this material is permitted.  Permission from IEEE must be obtained for all other uses, in any current or future media, including reprinting/republishing this material for advertising or promotional purposes, creating new collective works, for resale or redistribution to servers or lists, or reuse of any copyrighted component of this work in other works.}
\newcommand\copyrightnotice{%
	\begin{tikzpicture}[remember picture,overlay]
	\node[anchor=south,yshift=10pt] at (current page.south) {\fbox{\parbox{\dimexpr\textwidth-\fboxsep-\fboxrule\relax}{\copyrighttext}}};
	\end{tikzpicture}%
}
\begin{document}

\maketitle
\copyrightnotice
\thispagestyle{empty}
\pagestyle{empty}

\begin{abstract}
A key capability for autonomous underground mining vehicles is real-time accurate localisation. While significant progress has been made, currently deployed systems have several limitations ranging from dependence on costly additional infrastructure to failure of both visual and range-sensor-based techniques in highly aliased or visually challenging environments. In our previous work, we presented a lightweight coarse vision-based localisation system that could map and then localise to within a few metres in an underground mining environment. However, this level of precision is insufficient for providing a cheaper, more reliable vision-based automation alternative to current range sensor-based systems. Here we present a new precision localisation system dubbed ``\sysname{}'', which learns a neural-network-based pixel sampling strategy for estimating homographies based on ceiling-facing cameras without requiring any manual labelling. This new system runs in real time on limited computation resource and is demonstrated on two different underground mine sites, achieving real time performance at $\sim$\bestFrameRate{} frames per second and a much improved average localisation error of $\sim$\bestAccuracy{} metre. 

\end{abstract}

\section{INTRODUCTION}
Real-time high-accuracy localisation for autonomous vehicles in underground mine sites is challenging due to a lack of GPS, severe lighting changes, dust and environment ambiguity. As the mining industry seeks to become more efficient, companies are looking for more economical technology that will enable less lucrative secondary mining resources to be feasibly mined. One consequence of this for navigating autonomous mine vehicles is that infrastructure-based techniques are less feasible, while range-based sensors are often expensive and have been reported to struggle in geometrically aliased environments such as long uniform tunnels. Low-cost vision-based localisation technologies are among the most promising alternatives for overcoming these limitations.

Among vision-based localisation methods, the state-of-the-art general-purpose SLAM (Simultaneous Localisation and Mapping) algorithm ORB-SLAM~\cite{murAcceptedTRO2015} has been shown to perform unsatisfactorily in underground mine site environments~\cite{Jacobson2018}. Our previous work~\cite{Zeng2017,Jacobson2018} on coarse localisation based on whole-image matching has a demonstrated localisation accuracy out-performing a state-of-the-art deep learning approach \cite{sunderhauf2015performance} with a mean localisation error of a few metres~\cite{Jacobson2018}. Because it localises to the nearest node in the database, its accuracy is limited to the resolution of the node separation in the map. The goal of the research presented in this paper is to build on the previously presented coarse localisation system to enable a higher degree of precision with the eventual aim of enabling reliable, vision-only autonomous control of underground mining vehicles.

\begin{figure}[tp]
	\centering
	\begin{subfigure}[b]{0.21\textwidth}
		\centering
		\includegraphics[width=\textwidth]{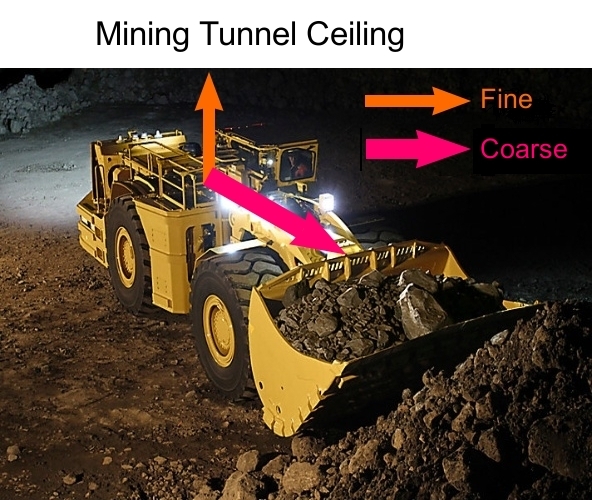}
		\caption{}
		\label{fig:truck_image}
	\end{subfigure}
	\begin{subfigure}[b]{0.11\textwidth}
		\centering
		\includegraphics[width=\textwidth]{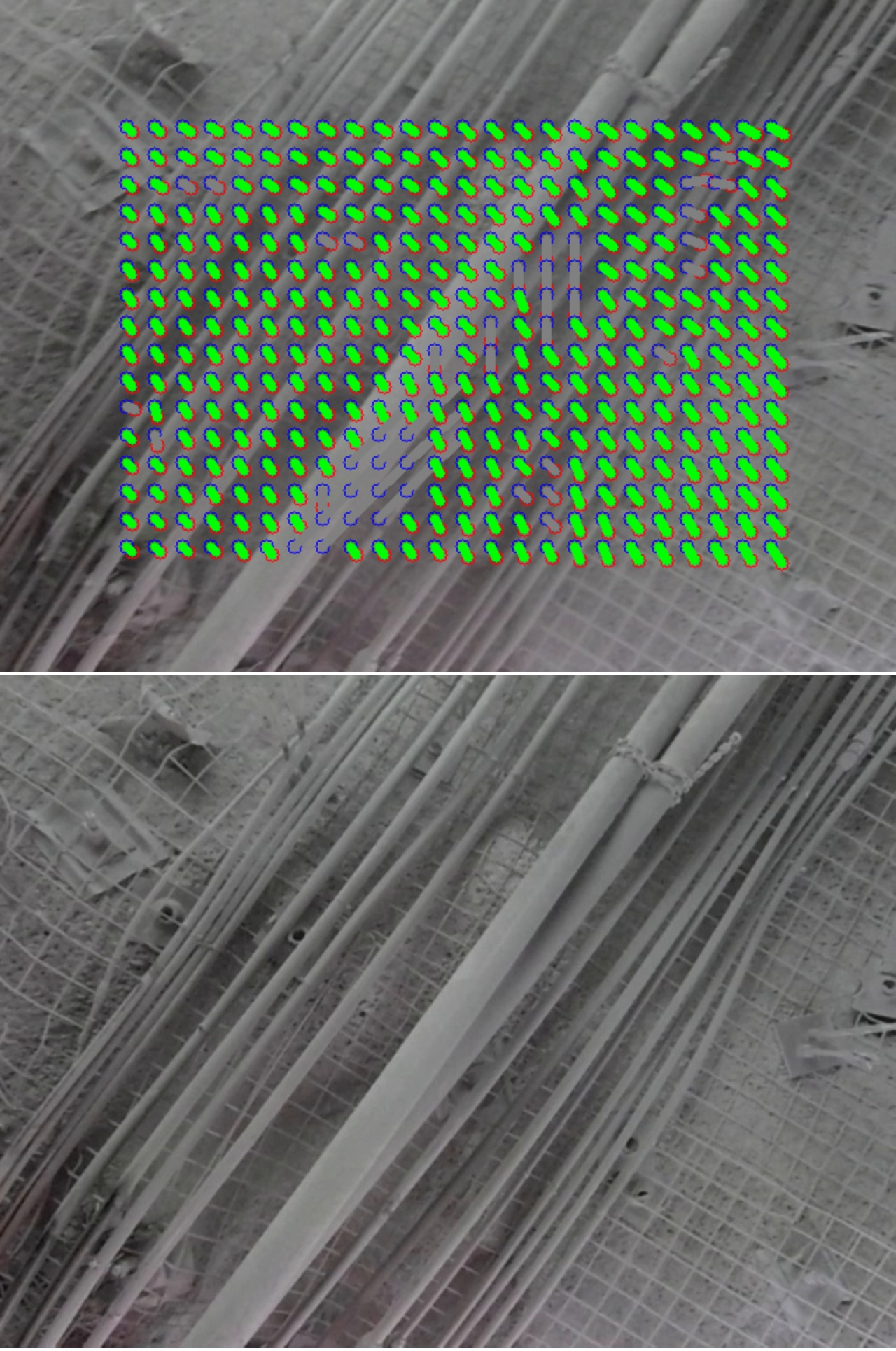}
		\caption{}
		\label{fig:intro_0}
	\end{subfigure}
	~
	\begin{subfigure}[b]{0.11\textwidth}
		\centering
		\includegraphics[width=\textwidth]{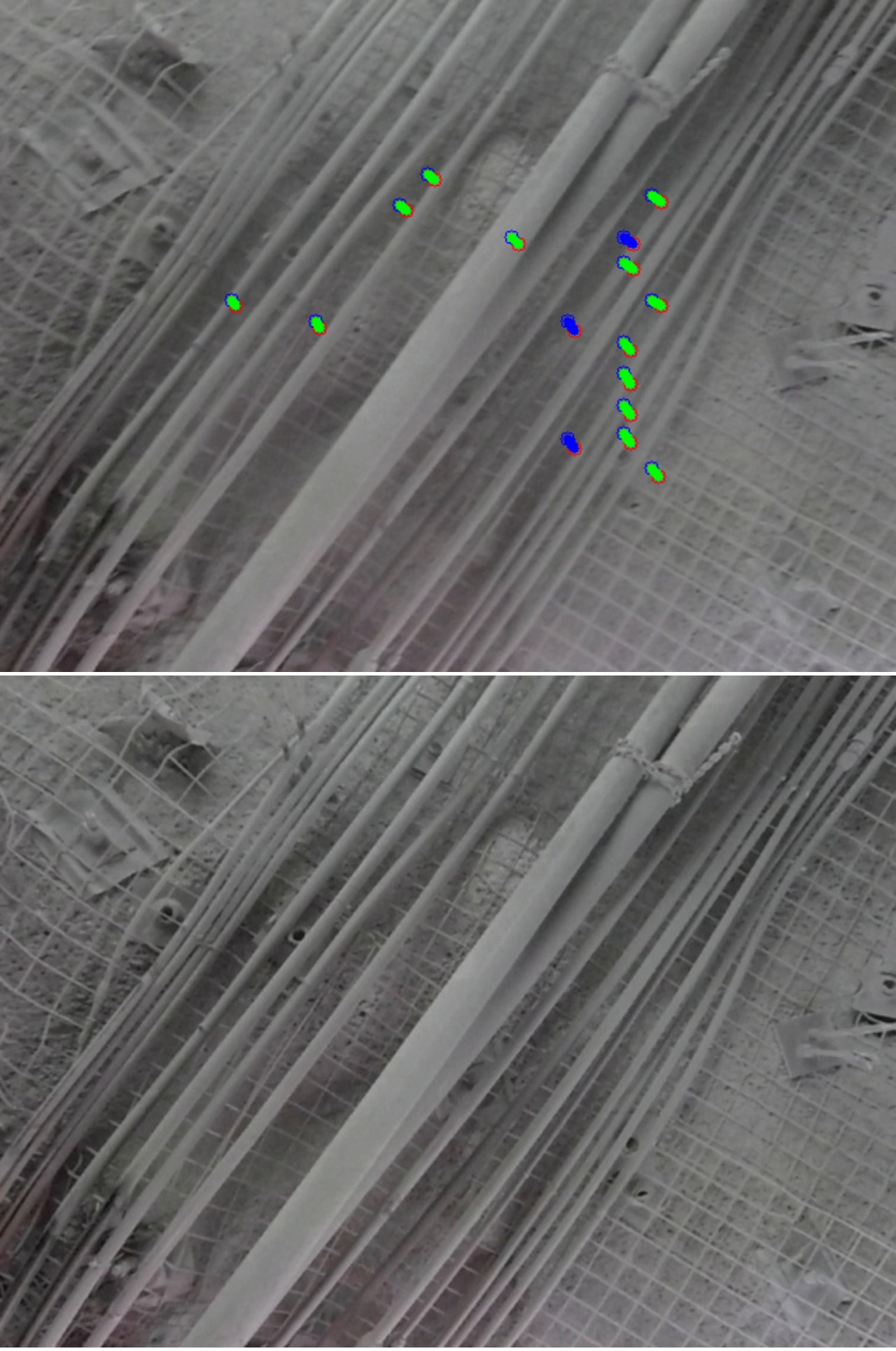}
		\caption{}
		\label{fig:intro_1}
	\end{subfigure}
	~
	\begin{subfigure}[b]{0.1\textwidth}
		\centering
		\includegraphics[width=\textwidth]{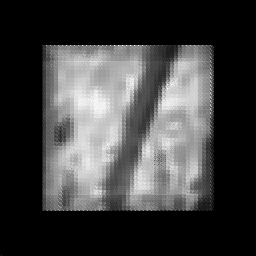}
		\caption{}
		\label{fig:sample_image.good_sample_points_heat_map.cannginton}
	\end{subfigure}
	~
	\begin{subfigure}[b]{0.1\textwidth}
		\centering
		\includegraphics[width=\textwidth]{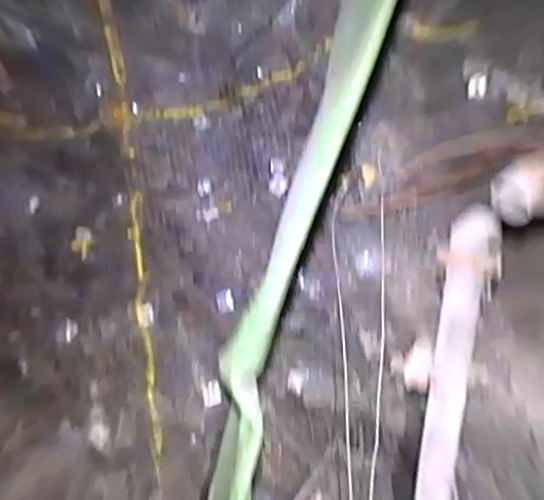}
		\caption{}
		\label{fig:sample_image.original_image.cannington}
	\end{subfigure}
	~
	\begin{subfigure}[b]{0.1\textwidth}
		\centering
		\includegraphics[width=\textwidth]{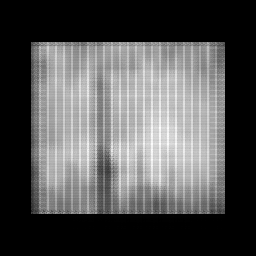}
		\caption{}
		\label{fig:sample_image.good_sample_points_heat_map.cadia}
	\end{subfigure}
	~
	\begin{subfigure}[b]{0.1\textwidth}
		\centering
		\includegraphics[width=\textwidth]{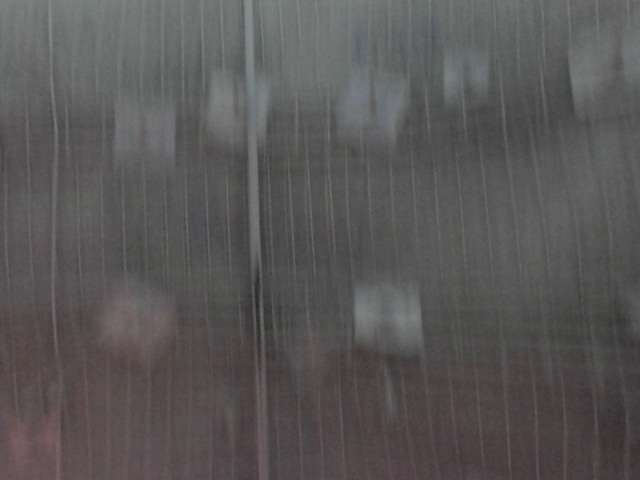}
		\caption{}
		\label{fig:sample_image.original_image.cadia}
	\end{subfigure}

	\caption{(a) The proposed system consists of a coarse localisation stage using a forward-facing camera (pink arrow) and a refinement stage, \sysname{}, using an upward-facing one (orange arrow). (b)(c) Examples of ``optical flow'' between a query image (top) and a nearby reference image (bottom) when (b) a regular grid is used, and (c) an FCN (Fully Convolutional Network) is used with \sysname{}. Note the significant reduction in number of sample points from (b) to (c). Inlier optical flow vectors are coloured in green. (d)(f) Sample point quality heat maps generated by the FCN in \sysname{} for the original images in (e)(g), respectively.}
	\label{fig:introduction}
	\vspace{-20pt}
\end{figure}

Developing localisation for underground autonomous vehicles presents some challenges and opportunities. Sensing and hardware capabilities are limited as sensors must be toughened, severely restricting the use of recent hardware and limiting the deployment of computationally intensive algorithms including full size deep learning architectures. There are also limitations in the practical amount of training data that can be obtained from a site. Naive deployment of full 6DOF SLAM systems (e.g. \cite{Newman2009}) is not necessary as there are a range of constraints that can be applied: the pitch and roll variations of the vehicle (therefore the camera) relative to the tunnel can be assumed to be limited, as is the variation in the height of the ceiling. Even allowing for the occasional three-dimensional structures such as wind pipes, the ceiling of mine tunnels is mostly planar. This offers an opportunity to significantly reduce computation, since theoretically as few as four point-correspondences are required for planar homography estimation. Furthermore, the ceiling-facing camera is less affected by dust and lighting from other vehicles.

The paper makes the following contributions:
\begin{itemize}
	\item A new vision-only localisation system, designed for underground mine environments, which takes coarse localisation results and refines them through rapid quasi-planar surface homography estimation.
	\item An efficient neural-network-based sample point selector that generates quality heat maps of candidate points for effective pixel-correspondence calculations, and an associated off-line training process that does not require manual dataset labelling.
	\item Demonstration of new levels of vision-only localisation accuracy in two new challenging underground mine site datasets.
\end{itemize}

The paper proceeds as follows. Section \ref{section:literature_review} reviews previous work on robust localisation algorithms and various saliency generation methods used as preprocessing filters for image matchers. Section \ref{section:approach} provides a detailed description of the proposed localisation system. Section \ref{section:experiments} describes experimental settings including the datasets and our method to build the evaluation benchmark, with the results presented in Section \ref{section:results} followed by the conclusion in Section \ref{section:conclusion}. 

\section{Literature Review} \label{section:literature_review}
\subsection{LIDAR-based Localisation Methods}
Laser scanners (LIDARs) can provide metric position estimations when there are ample features across the scanned angle span, but laser-scanner-based localisation systems \cite{Cole2006,magnusson2009appearance,Sprunk2013,Bosse2007} can easily get lost in long tunnels, which are ubiquitous in underground mines, as the scanned point clouds appear confusingly similar along the tunnel. This problem is uncommon in environments such as typical rooms and warehouses because the shape of enclosing walls provides salient variations across the scanned angle span. In a long tunnel, a LIDAR essentially becomes one dimensional - it only knows its distance to the walls but has no idea about how far it has travelled along them. Moreover, in the areas of the mine where there are more features, such as draw points, there could be objects like metal meshes that could confuse localisation methods based on 2D laser scanners, because the returns from the mesh may also form occupied space that could be misinterpreted as a wall to align scans with. Therefore, due to the current limitations of LIDAR-based methods, we choose to exploit vision-based place recognition methods to do localisation, for enhanced global robustness.

\subsection{Vision-Based Methods}
Traditional feature-based place recognition algorithms such as FAB-MAP \cite{cummins2008fab,Cummins2009} work poorly in mine-tunnel environments due to severe visual aliasing. SeqSLAM  \cite{Milford2012} and many other SLAM frameworks \cite{Milford2004a,Maddern,Glover2010} are less sensitive but require external sources like GPS or wheel odometry to provide metric information. As demonstrated by our previous results, the coarse localisation unit -  \coarseLocalisationUnitName{} - is able to produce better maps than ORB-SLAM \cite{murAcceptedTRO2015} with 2.5 times smaller localisation error. Nevertheless, a higher localisation accuracy is desirable to better assist the automated control of vehicle pose during various activities such as digging, dumping and driving.

\begin{figure}[tp]
	\centering
	\begin{subfigure}[b]{0.14\textwidth}
		\centering
		\includegraphics[width=\textwidth]{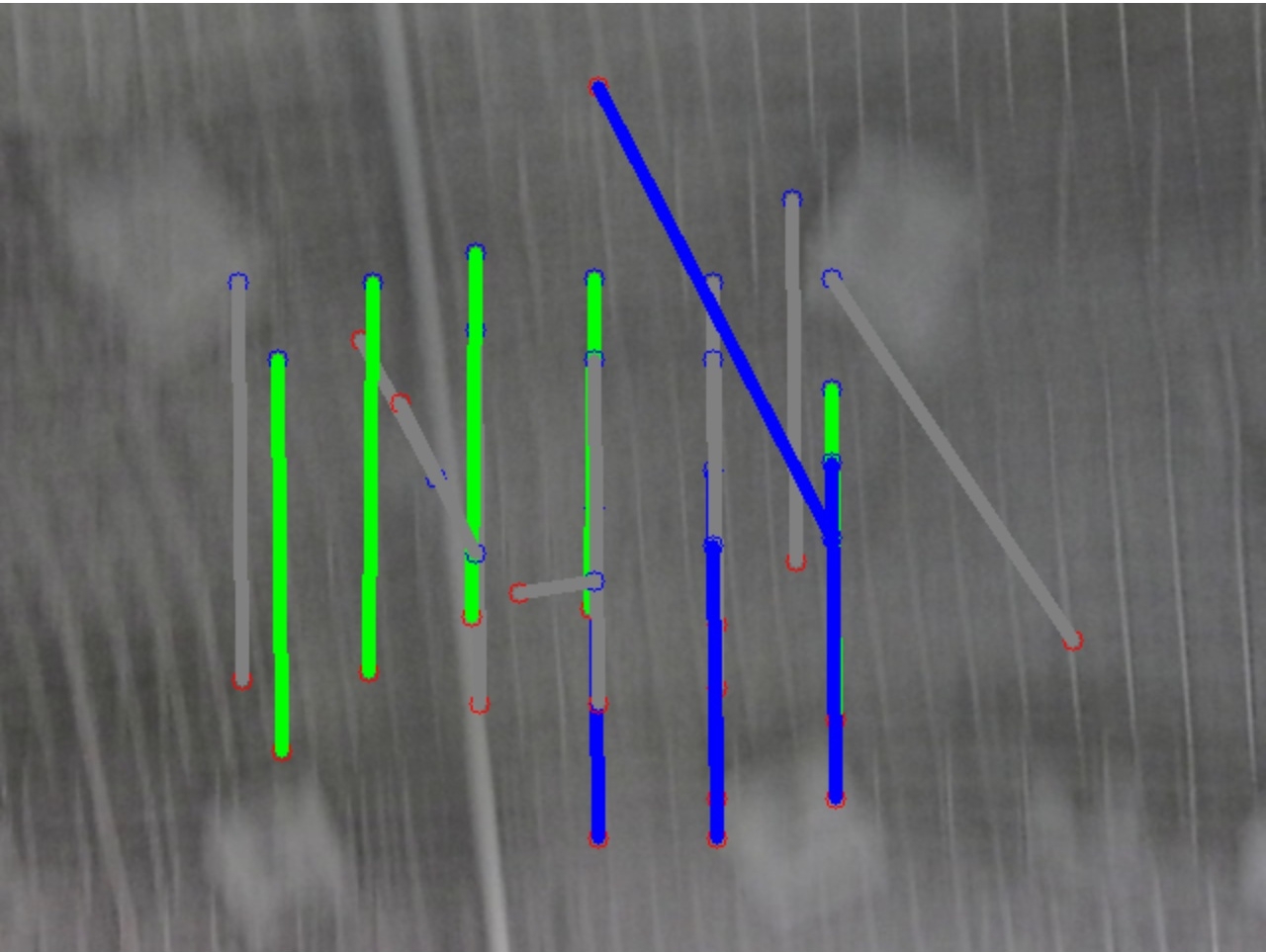}
		\caption{}
		\label{fig:intro_large_optical_flow_a}
	\end{subfigure}
	~
	\begin{subfigure}[b]{0.14\textwidth}
		\centering
		\includegraphics[width=\textwidth]{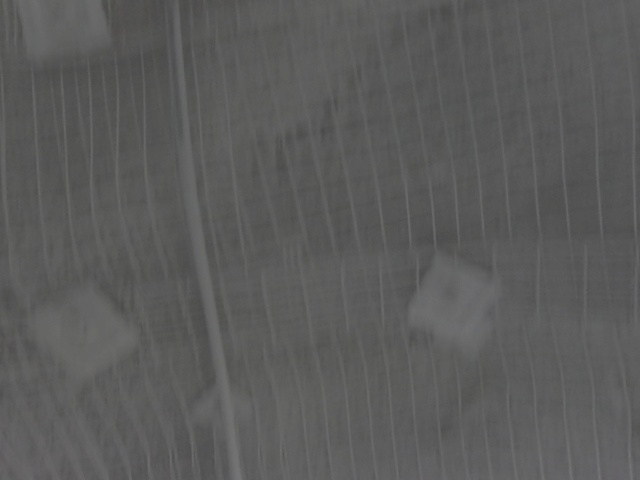}
		\caption{}
		\label{fig:intro_large_optical_flow_b}
	\end{subfigure}
	~
	\begin{subfigure}[b]{0.14\textwidth}
		\centering
		\includegraphics[width=\textwidth]{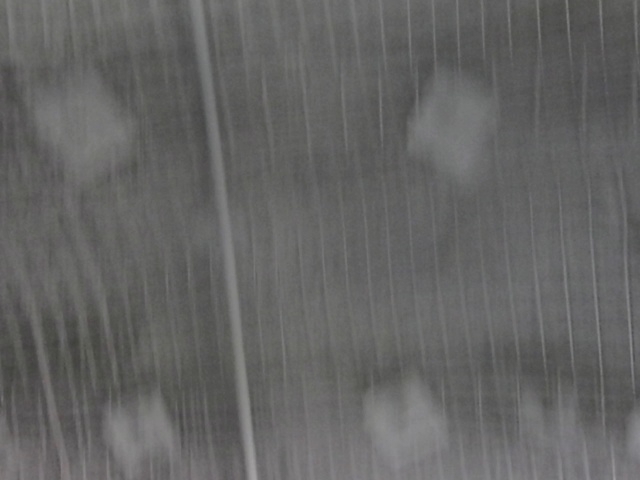}
		\caption{}
		\label{fig:intro_large_optical_flow_c}
	\end{subfigure}
	\caption{(a) Displacement (optical flow) between matched images - shown in (b)(c) - could be large. Blue optical flow vectors were rejected by the Pixel Correspondence Matcher, the rest were RANSAC filtered, with inliers and outliers shown in green and gray, respectively.}
	\label{fig:intro_large_optical_flow}
	\vspace{-20pt}
\end{figure}

Given the range of uncertainty of the coarse localisation results and the sparse density of reference images sampled across the mine, the translation between reference and query images can be quite significant comparing to the captured range, even when a wide Field Of View (FOV) camera is used, because the walls and ceiling of the tunnel are usually a short distance away from the camera. As a result, the matched point pair, if it exists, can be a large distance apart (Fig.~\ref{fig:intro_large_optical_flow}), under limited frame-rate constraints. Although we still refer to this translation vector as ``optical flow'', traditional optical flow algorithms \cite{lucas1981iterative,shi1993good} typically assume small displacement \cite{lepetit2005monocular} and are not suitable for our application. I2-S2 \cite{quteprints125531} has been proposed to extract homographies between query and candidate reference images for pixels at predefined image locations. Different saliency generators \cite{Milford2014b,Zeng2017} have been proposed for sample point or patch filtering, however, they are based on pre-determined metrics of pixel intensities and do not adapt automatically to a different context.

\subsection{Deep Convolutional Networks}
Deep convolutional networks \cite{krizhevsky2012imagenet} have been proven to be successful in place recognition \cite{sunderhauf2015performance,arandjelovic2016netvlad}, image classification and semantic segmentation \cite{girshick2015fast,ren2015faster}. However, there is no direct metric information output from these methods. Deep learning based methods \cite{dosovitskiy2015flownet,weinzaepfel2013deepflow,Wulff:CVPR:2017} have also been used to analyse large optical flows, among which FCN-based pixel labelling  \cite{long2015fully,noh2015learning} is suitable for our application of sample point selection, and an FCN similar to \cite{long2015fully} is used in this paper. In the next section, our precise localisation unit ``\sysname{}'' will be described.

\section{Approach} \label{section:approach}
The more precise localisation unit takes in a query image, a coarse localisation result, and has access to a database of images with known camera poses. This database can be collected with a single camera or an array of cameras during the surveying process accompanying the construction of a mine. The associated poses can be obtained via surveying tools and recorded alongside the image frames. Based on the coarse localisation result, relevant reference images in the database are cross-examined with the query image. Since the ceiling of mine tunnels provide quasi-planar surfaces to allow homography calculation based on only a handful of points, the cameras used in the precise localisation unit looks up towards the ceiling; in addition, the pose estimation requires a ``look up'' in the database to find the reference camera pose, hence the name ``\sysname{}''.

\begin{figure}[htp]
	\centering
	\includegraphics[width=0.4\textwidth]{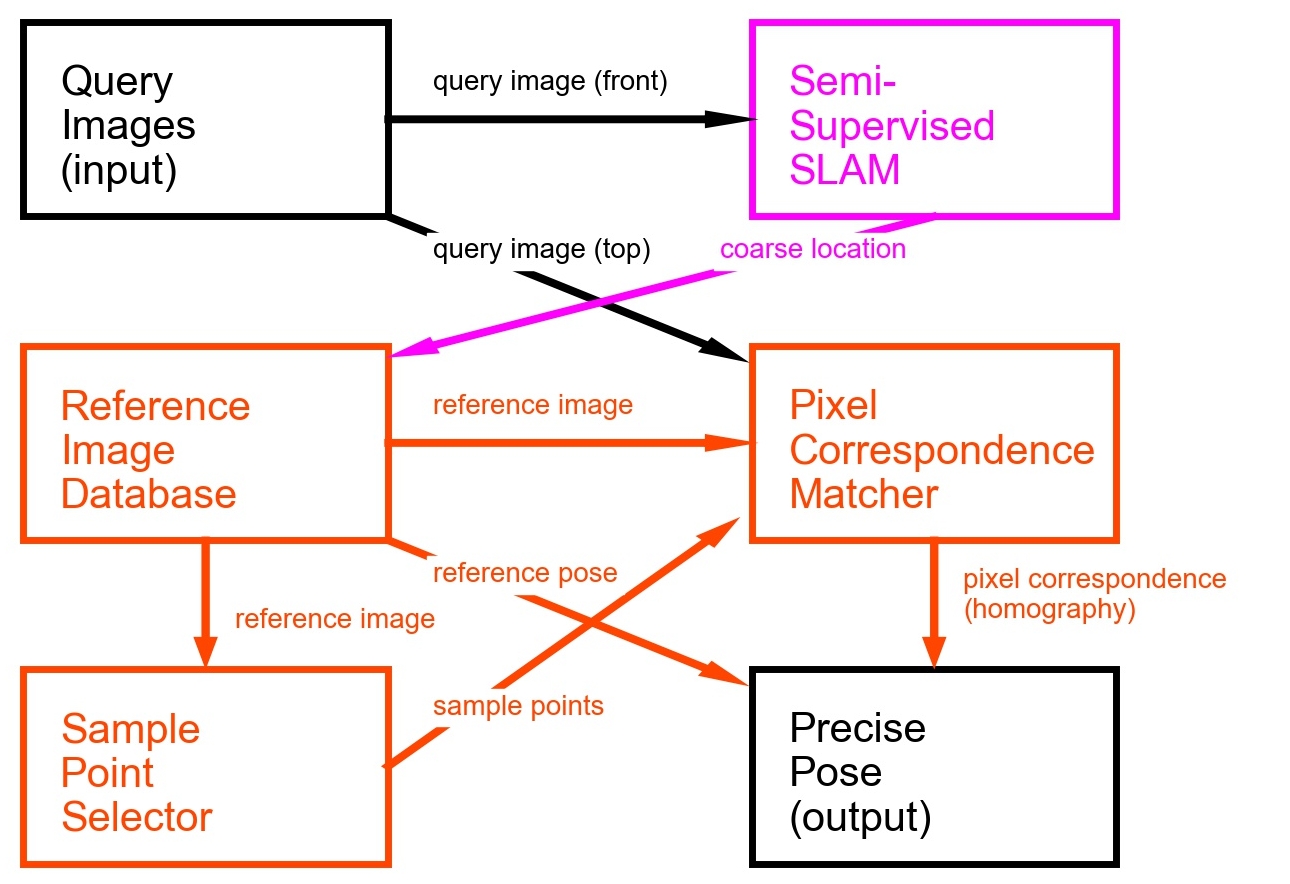}
	\caption{Schematic diagram of the underground localisation system showing the query images (black), the coarse localisation unit (pink) and the precise localisation unit \sysname{} (orange).}
	\label{fig:schematic_diagram}
	\vspace{-20pt}
\end{figure}

\subsection{Pixel Correspondence Matcher}
\begin{figure}[htp]
	\centering
	\begin{subfigure}[b]{0.22\textwidth}
		\centering
		\includegraphics[width=\textwidth]{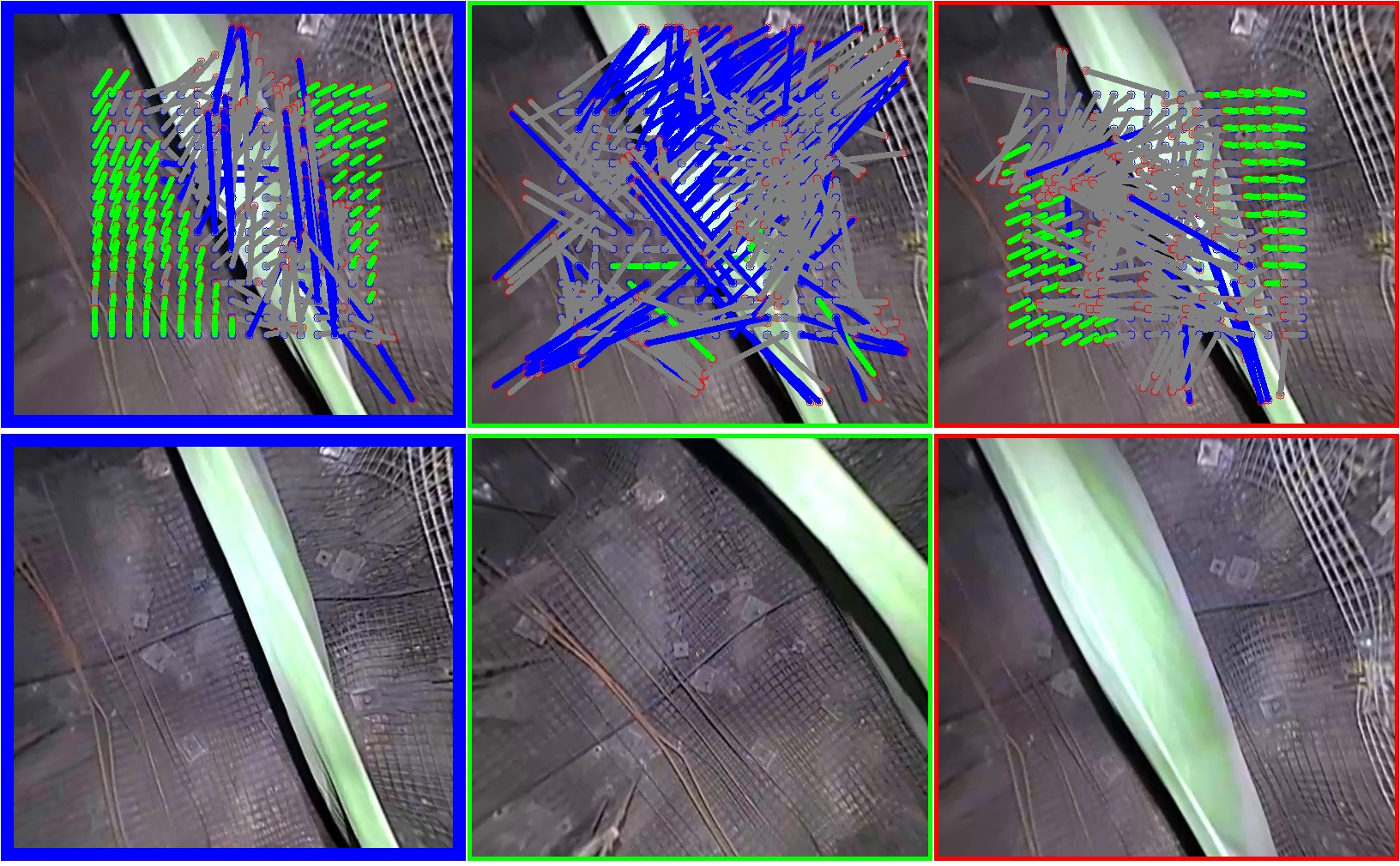}
		\caption{}
		\label{fig:sample_quality_example.0}
	\end{subfigure}
	~
	\begin{subfigure}[b]{0.22\textwidth}
		\centering
		\includegraphics[width=\textwidth]{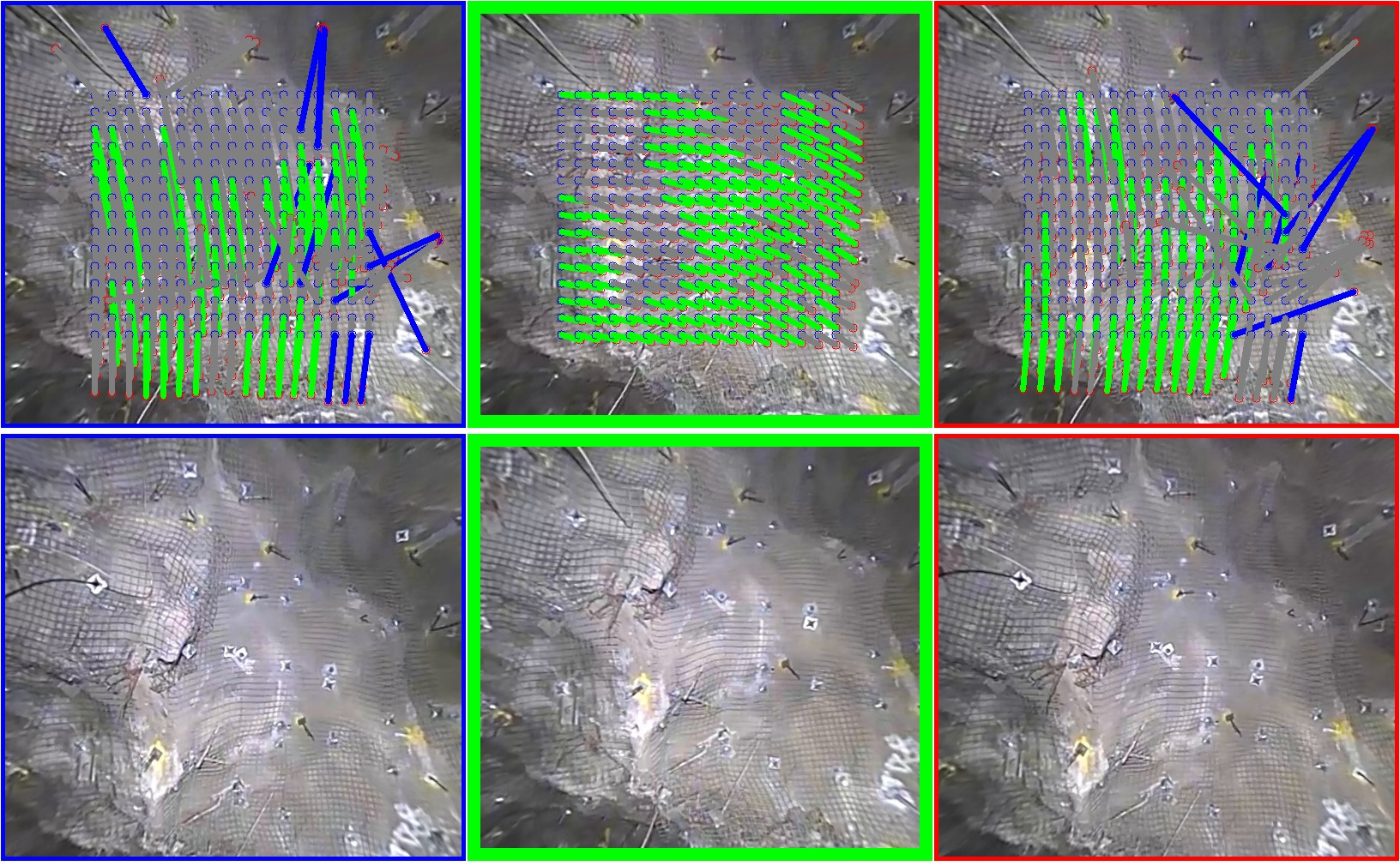}
		\caption{}
		\label{fig:sample_quality_example.1}
	\end{subfigure}
	\caption{Two examples of ``optical flow'' between a query image (top row) and a set of reference images (bottom row), featuring ``optical flow'' outliers caused by (a) 3D objects on the ceiling, and (b) uneven rock surfaces.}
	\label{fig:sample_quality_example}
	\vspace{-10pt}
\end{figure}

The Pixel Correspondence Matcher (Fig.~\ref{fig:schematic_diagram}) is used to find the most-likely corresponding pixel in a query image for a selected pixel in the reference image. It takes an $l_{patch}$-sized reference patch centred at the selected reference pixel, and generates a search neighbourhood in the query image, which is an $L_{SR} \times L_{SR}$ sized square centred at the same pixel coordinates as the selected reference pixel. It then compares the reference patch to a set of candidate patches centred at every pixel in search neighbourhood. The best match candidate pixel in terms of Sum of Absolute Difference (SAD) score is reported. The process is visualised as colour-coded ``optical flows'' in Fig.~\ref{fig:sample_quality_example}.

The processing time for each pair of reference and query images is proportional to the number of sampled pixels for which the pixel correspondence is to be found. Under the small pitch and roll assumption, most likely the inlier vectors from the output of a RANSAC \cite{fischler1981random} filter are similar in direction and magnitude (Fig.~\ref{fig:intro_0}). If we can identify such inliers in advance and only find pixel correspondence for them, the computation time could be reduced. As for the outliers, they are excluded from the homography calculation anyway, it would be better if they were not sampled in the first place. In this paper we used an off-the-shelf neural network (VGG16-based FCN) to produce sampling point qualities. 

\subsection{Sample Point Selector}
As can be seen from Fig.~\ref{fig:sample_quality_example}, the outlier sample points that produce inconsistent optical flow vectors are most likely on 3D objects (Fig.~\ref{fig:sample_quality_example.0}) or uneven rock surfaces (Fig.~\ref{fig:sample_quality_example.1}) on the ceiling. However, rather than defining rigid rules for classification, such as ``avoid long wires, pipes and strong lights'', more general and adaptive qualification criteria are desirable. This is because in some situations, certain objects may provide high-quality sample points for template matching, but they may not work well in other cases - the semantics of the features affect their quality as a sample point, involving contextual information many pixels away from them. Support Vector Machines~\cite{suykens1999least} are usually effective binary classifiers but they are limited to local information around the sample point. A neural network architecture that incorporates more holistic information is preferred. 

Although feature-based methods are susceptible to visual aliasing in our underground localisation application, they may work well for sample point quality generation because visual aliasing is not a problem for this task. We implemented an FCN similar to the one described in \cite{noh2015learning}. The query image is fed into the convolutional layers of a VGG16~\cite{Simonyan14c} network pre-trained with the ImageNet dataset \cite{deng2009imagenet}, the output of which goes into a $1 \times 1$ convolution and three up-sampling layers, with skip connections to layer3 and layer4 of the original VGG16. The output of the network is a heat map of sample point quality (Figs.~\ref{fig:sample_image.good_sample_points_heat_map.cannginton} and \ref{fig:sample_image.good_sample_points_heat_map.cadia}), according to which the sample points are selected (Fig.~\ref{fig:schematic_diagram}).
The training dataset of the FCN is generated by applying the Pixel Correspondence Matcher in \sysname{} to a training image dataset, processing points densely sampled on a regular grid, as shown in Figs.~\ref{fig:intro_0} and \ref{fig:sample_quality_example}. RANSAC is used to classify the sampled points into inliers (coloured coded green) and outliers (colour coded gray), according to the optical flow vector obtained on that sample point. The FCN is then trained using this labelled data. The loss function is defined as proportional to the total number of misclassified sample points for the training images.  The output of the sample point selector is a heat map of quality for all candidate pixels. After the FCN is trained, all the reference images are processed with it and corresponding sample quality heat maps are generated alongside the reference image database. The training and classification processes are completed off-line, therefore they neither take up on-line run time nor require a GPU in the localisation system. At run time, it is up to the pixel correspondence matcher to decide how this heat map should be used. 

\subsection{Homography Estimator}
The set of ``optical flow'' vectors calculated by the pixel correspondence matcher from all selected sample points are used to compute a $3 \times 3$ homography matrix that relates the pose of the query and reference images. Although multiple solutions exist for the homography matrix, it is not hard to identify the one that makes physical sense by choosing the solution that gives the smaller pitch and roll. Before the homography is found, there is an optional RANSAC filtering if the number of sampled points is greater than 10.

\subsection{Determination of Scaling Constant}
The above homography estimation process can be done with multiple reference images (each column in Fig.~\ref{fig:sample_quality_example}(a)(b)). If there are more than one reference images for which good matches are found, it is possible to estimate the constant that converts the distance from pixel to metric space, using the assumption that the scaling constant should be similar for both homography relations. If only one reference image is used for faster processing, it is also possible to use a pre-determined constant for this conversion under the assumption of small variations in the ceiling height.

\subsection{Integration with the Coarse Localisation Unit}
Currently, the interface between \sysname{} and \coarseLocalisationUnitName{} is simply the time stamp of the database image that is considered a match. \sysname{} will fetch the reference images from the ceiling-facing camera that were taken most closely in time to the matched database image from the forward facing camera. A refined location is estimated by \sysname{} using this reference image and the system then decides whether this refinement should be applied. Two filters are applied. The refinement is deemed not reliable if 1) the percentage of inliers after the RANSAC filtering is lower than a threshold $N_{th}$ or 2) if the $x$ or $y$ translation from the reference pose extracted from the homography is larger than a threshold $d_{th}$. These could happen if the coarse localisation result is incorrect, or the relative displacement is larger than the search range. The system will simply fall back to the coarse localisation result when \sysname{} is not confident. Apart from the above interface, the coarse and fine localisation units are highly independent and can be optimised separately. Next we describe the experiments we have done to evaluate the performance of \sysname{} system.

\section{Experiments} \label{section:experiments}

In order to evaluate the precise localisation system, coarse localisation needs to be performed first. Based on a map in which the reference poses are defined, the coarse localisation system, \coarseLocalisationUnitName{} \cite{Jacobson2018}, takes in images with known locations and constructs an internal database according to their associated locations, grouping images taken at adjacent places to the same node and saves them in a database. When sequences of query images arrive, it compares query images to reference images in the database, and generates a confusion matrix corresponding to the sequence of query images. Using the confusion matrix, \sysname{} was then run to output metric location results in the map.

To evaluate the refinement achieved by \sysname{}, the localisation results corresponding to the confusion matrix were also generated by disabling \sysname{} and directly outputting the reference poses corresponding to the time stamp of the matched reference image. The frames for which \coarseLocalisationUnitName{} generated localisation errors that were greater than 10 metres, for which a refinement is hardly possible, were excluded from the evaluation. Next we describe the real-world datasets collected to do such evaluation and how the maps, reference poses and benchmark localisation results were obtained.

\subsection{Datasets} \label{subsection:datasets}

In order to evaluate the localisation accuracy, a different localisation system that can generate benchmark localisation results that are at least locally accurate must be applicable to the datasets. If the datasets contain many draw points and junctions but few long stretches of tunnels, algorithms based on laser scan matching can be used for benchmarking. Based on such criteria, the following datasets were collected. 

\begin{enumerate}

\item Mine A dataset:
This dataset includes nine traverses of a heavy vehicle in two connected tunnels of an underground mine (Fig.~\ref{fig:slam_map.cadia}). Four of the traverses are used to build the map and the reference image database, the other five are used as localisation query. This is the same dataset used in \cite{Jacobson2018}. 
\item Mine B dataset:
The majority of the optical flows between images in the Mine A dataset are along the travelling direction of the vehicle. On the other hand, \sysname{} does not constrain the optical flow search along one direction. To study the generality of \sysname{}, a second dataset was collected in a different mine, featuring four traverses of a light vehicle in a mine tunnel (Fig.~\ref{fig:slam_map.cannington}). Traverse Middle(M): the light vehicle was driven along the centre of the tunnel. Traverse Left(L) and Right(R): the light vehicle was driven close to the left and right wall, respectively. Traverse Zigzag(Z): the light vehicle was driven deliberately in a zigzag motion. Traverse M was used to build the SLAM map; Traverses L, M and R were used to build the reference image database; Traverse Z was used as the localisation query. In this way, the query images in this dataset can have optical flows in various directions w.r.t the references.

Altogether the two datasets contain \datasetFrameNumber{} data frames over \datasetTimeSpan{} seconds of $\sim$\datasetLength{} kilometre traverses (average vehicle speed $\sim$35 km/h). These datasets are particular challenging due to the affluence of heavily aliased patterns on multiple scales.
\end{enumerate}

\subsection{Map Building and Reference Poses}
\begin{figure}[htp]
	\centering
	\begin{subfigure}[b]{0.45\textwidth}
		\centering
		\includegraphics[width=\textwidth]{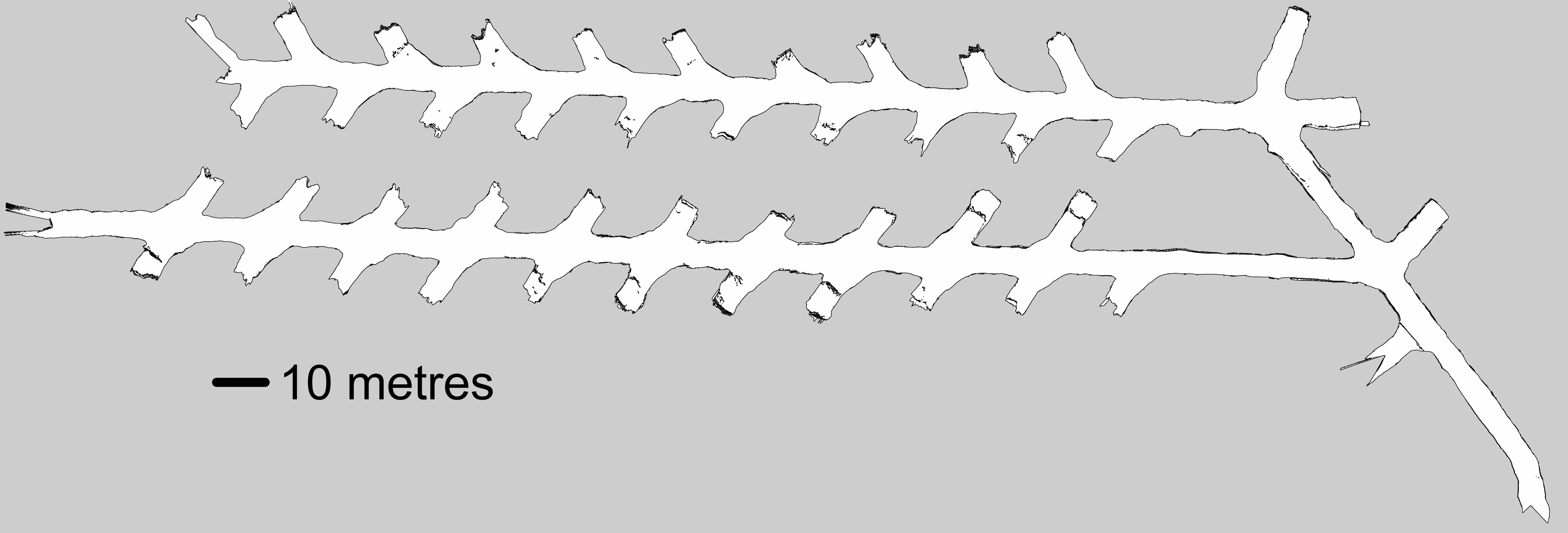}
		\caption{SLAM Map of Mine A, built by Cartographer \cite{45466}.}
		\label{fig:slam_map.cadia}
	\end{subfigure}
	\begin{subfigure}[b]{0.45\textwidth}
		\centering
		\includegraphics[width=\textwidth]{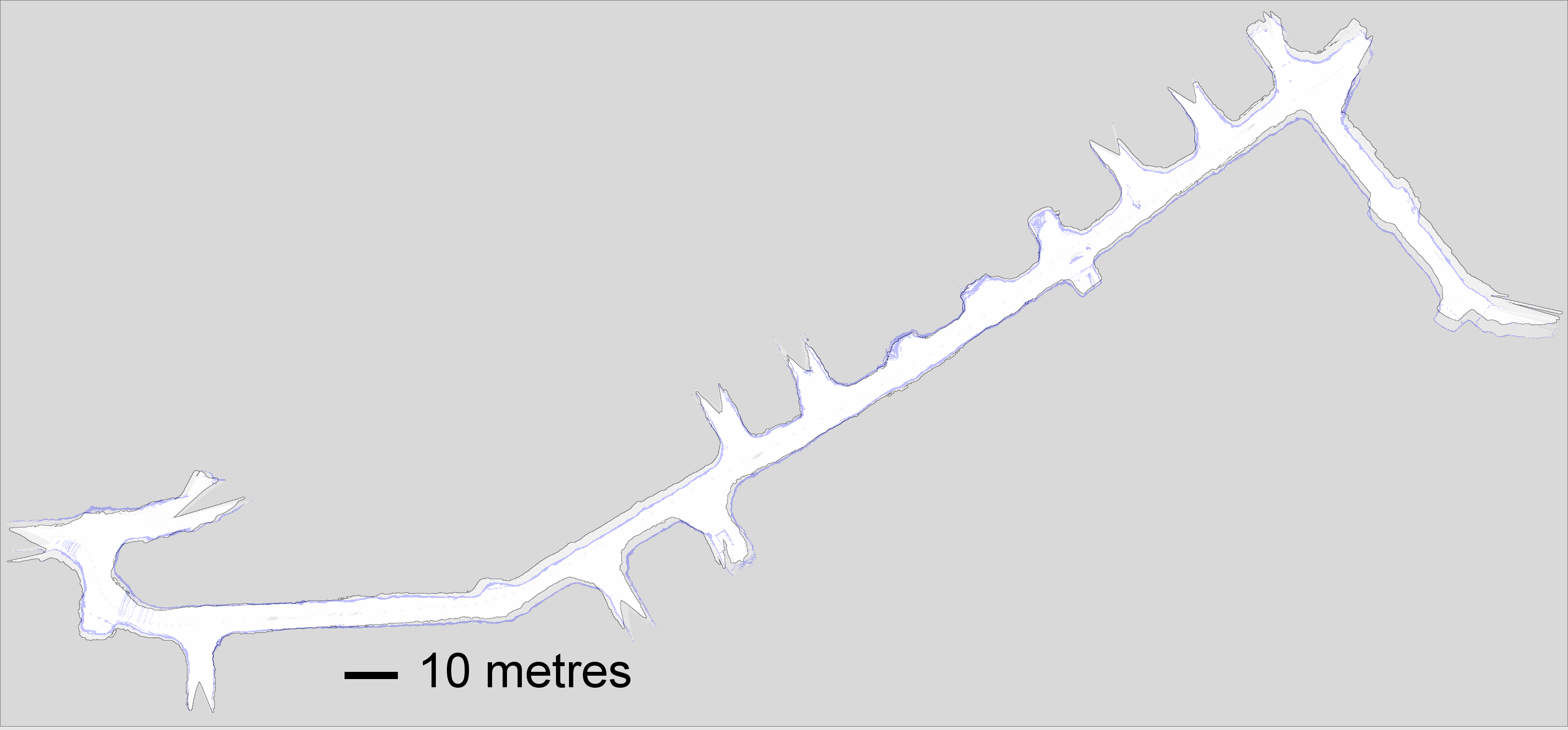}
		\caption{SLAM Map of Mine B. Blue: Hector-Mapping \cite{KohlbrecherMeyerStrykKlingaufFlexibleSlamSystem2011}; Black: Cartographer \cite{45466}.}
		\label{fig:slam_map.cannington}
	\end{subfigure}
	\caption{Maps built by the SLAM algorithms in \cite{KohlbrecherMeyerStrykKlingaufFlexibleSlamSystem2011,45466}. Note our system does not depend on these algorithms.}
	\label{fig:slam_map}
	\vspace{-10pt}
\end{figure}

\begin{enumerate}
\item Mine A:
The coarse localisation results were directly taken from \cite{Jacobson2018}. However, the metric locations from \cite{Jacobson2018} were based on an external Radio Telemetry System that is not accurate enough for evaluating the precise localisation system. A more precise occupancy grid map was required to generate the reference and benchmark poses.
The attempt to build such a map using Hector-Mapping \cite{KohlbrecherMeyerStrykKlingaufFlexibleSlamSystem2011} was unsuccessful since this dataset contains a few sections of long tunnels and metal meshes. Therefore, a different approach was used to build the map. First, four separate maps, one for each reference traverse were built using Cartographer \cite{45466}, then the four maps were manually aligned to form a large map, shown as the black occupancy grid in Fig.~\ref{fig:slam_map.cadia}. The manual assembly was necessary because the four traverses used for map building were not collected continuously in time and space. The reference poses were then obtained by running AMCL \cite{dellaert1999monte} on the stitched map subscribing to the ``ROS tf frames'' \cite{quigley2009ros} published by Cartographer.
\item Mine B:
The map of mine tunnel was successfully built using Hector-Mapping, shown as the blue occupancy grid in Fig.~\ref{fig:slam_map.cannington}. The camera poses of the reference images were obtained by running AMCL on the map subscribing to the ``ROS tf frames'' published by Hector-Mapping. Unlike Mine A dataset, no coarse localisation results were available, so AMCL was used on the same map used to generate the locations associated with the images used in \coarseLocalisationUnitName{}.

It should be clarified that it is not necessary to obtain reference poses in this way. We obtained the reference poses using the laser scan data with occupancy grid map simply because this dataset was collected after the construction of the mine and we did not have surveying capabilities.

\end{enumerate}

\subsection{Localisation benchmark}
To calculate the localisation errors for evaluating different system settings, AMCL was run on the query traverses to produce the benchmark poses. During the AMCL runs, the poses of the vehicle and the laser scan results are visualised together with the maps (in Fig.~\ref{fig:slam_map}). Except for the beginning of each traverse, when AMCL is ``initializing likelihood field model with probabilities'', and a few times in the tunnel sections in the maps (Fig.~\ref{fig:slam_map}) where there are no draw points or junctions, the laser scans align with the map pretty well. Since the reference poses are built with the same maps in the same way, although the maps and the AMCL poses may not be globally accurate, the AMCL poses can be reasoned as locally reliable enough to be used for the local refinements presented in this paper, which are essentially relative pose transformations indifferent to absolute global coordinates.

Additionally, the global accuracy of the whole system is cross-verified with an independent algorithm on the Mine B dataset. The state-of-the-art SLAM algorithm - Cartographer \cite{45466}, not used for Mine B, was chosen to build a second set of map (black occupancy grid in Fig. \ref{fig:slam_map.cannington}). The two SLAM algorithms work under different principles: AMCL uses particle filters and Cartographer uses iterative optimisations of pose graph. Proper loop closure was achieved by both algorithms, which is non-trivial for such datasets. As shown in Fig.~\ref{fig:slam_map.cannington}, the difference between the two maps is within \GTAccuracy{} metres, indicating the accuracy of the AMCL poses in a more global sense.

\subsection{Comparison of FCN with Regular Grid}
The FCN was implemented with Tensorflow \cite{abadi2016tensorflow} in Python. It was trained with Stochastic Gradient Descent (SGD) with batch size of 8 (the maximum that can fit into an NVIDIA GeForce GTX 1080 GPU) and drop out rate of 50\%. Adam Optimiser and Softmax activation were used to generate the sample quality heat map. \sysname{} iteratively selects the best sample point (the one with highest heat map value), and apply a fixed reduction ratio $\rho$ to its $l_n$-sized neighbourhood in the heat map. It continues to pick the next best sample point until the required sample point number is reached. The FCN-based sample point selector was evaluated on the Mine A dataset in comparison with a regular grid sampling method. The regular grid contains 24 sample points (at the cost of more computation), whereas only the top 12 from the FCN-based sample point classifier were processed. All other parameters were kept the same. Selected frames of query images from query traverse 0 were used to train the FCN. After that the FCN generated sample point quality for all reference images in the database, which does not include any image the FCN was trained on. The FCN for Mine B dataset was trained on sub-sampled query frames, and classified sample points for the reference images.

\subsection{Parameters}
The parameters in Table \ref{tbl:parameters} were used to obtain the results in the next section.
\begin{table}[htp]
	\centering
	\caption{PARAMETER LIST}
	\begin{tabular}{r c c p{3.5cm}}
		\hline
		\toprule		
		Parameter & Value & Unit & Description \\
		\midrule
		$L_{SR}$ & 40 & pixels & Search range, Mine A\\
		$L_{SR}$ & 70 & pixels & Search range, Mine B\\
		$l_{patch}$ & 40 & pixels & Patch size, Mine A\\
		$l_{patch}$ & 60 & pixels & Patch size, Mine B\\
		$\rho$ & 0.5 &  & Factor multiplied to heat map value within neighbourhood of currently selected sample point\\
		$l_{n}$ & 10 & pixels & Neighbourhood size, Mine A\\
		$l_{n}$ & 20 & pixels & Neighbourhood size, Mine B\\
		$N_{th}$ & 60\% &  & Min. inlier percentage \\
		$d_{th}$ & 2 & metre & Max. displacement threshold  \\
		\bottomrule
		\hline
			\vspace{-20pt}
	\end{tabular}
	\label{tbl:parameters}
\end{table}


\section{Results} \label{section:results}
\subsection{Evaluation of the FCN}
The performance of the FCN in generating high-quality sample points was evaluated on test sets of images different from the training sets. The classification accuracy of the best sample point selected by the FCN was compared with that of a random point generator (representing the percentage of good sampling points in the ground truth). The percentage of correct classifications for test sets of Mine A dataset was $\sim$74\%, compared to $\sim$62\% from a random sampler; for Mine B dataset it was $\sim$41\% compared to $\sim$11\%.
\subsection{Localisation Results of \sysname{}}

%

\begin{figure}[htp]
	\centering
	
	\begin{subfigure}[b]{0.45\textwidth}
		\centering
		\includegraphics[width=\textwidth]{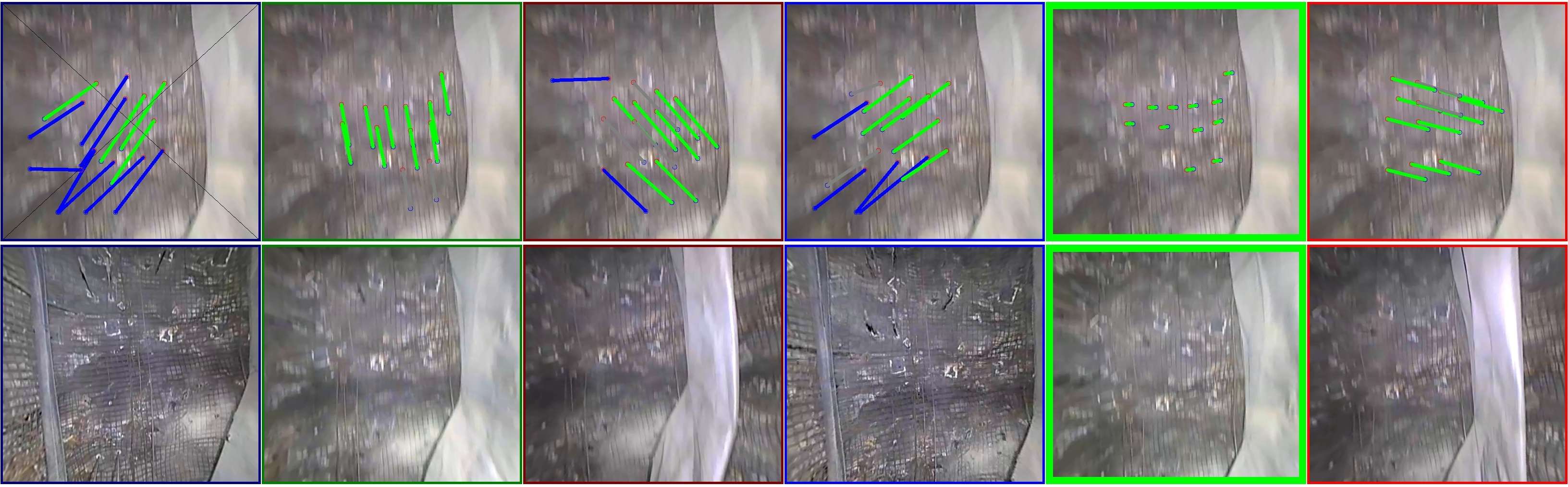}
		\caption{}
		\label{fig:localisation_result.cannington.sample_frame}
	\end{subfigure}
	~
	\begin{subfigure}[b]{0.14\textwidth}
		\includegraphics[width=\textwidth]{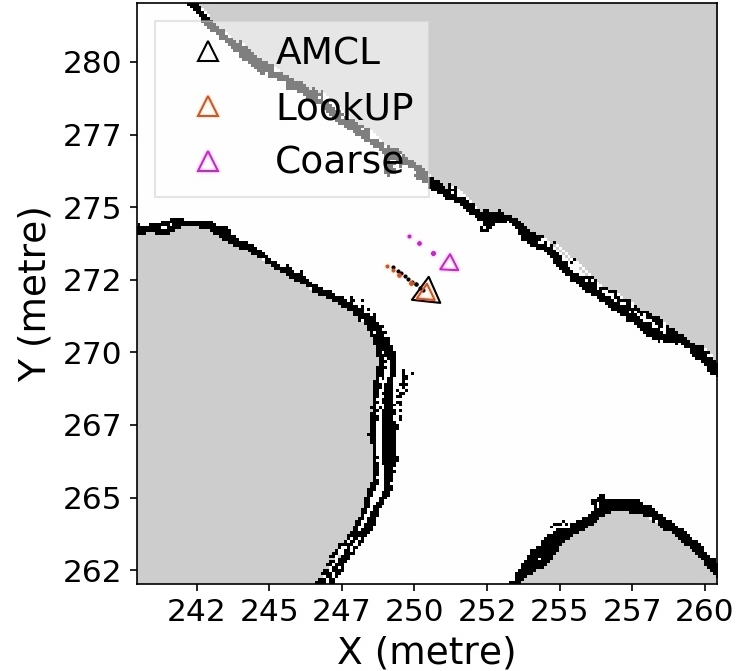}
		\caption{}
		\label{fig:localisation_result.canninton.pose_in_map_0}
	\end{subfigure}
	~
	\begin{subfigure}[b]{0.14\textwidth}
		\includegraphics[width=\textwidth]{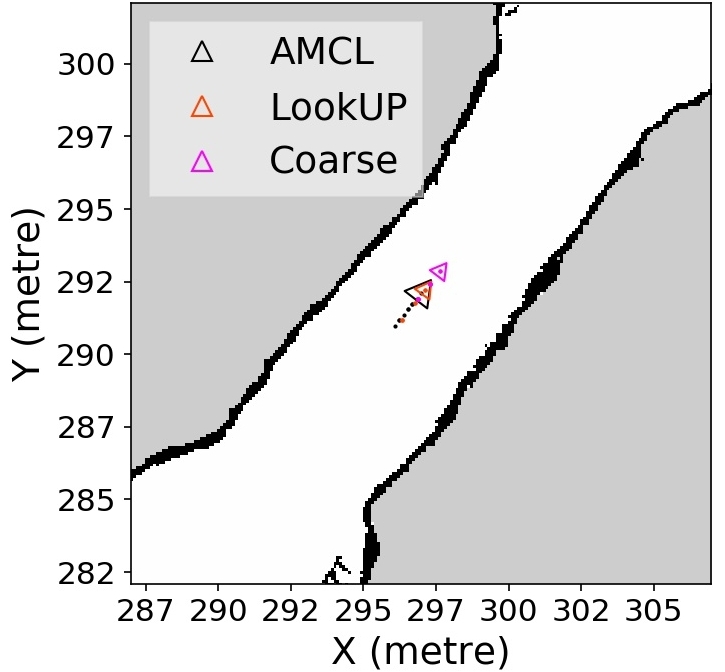}
		\caption{}
		\label{fig:localisation_result.canninton.pose_in_map_1}
	\end{subfigure}
	~
	\begin{subfigure}[b]{0.14\textwidth}
		\includegraphics[width=\textwidth]{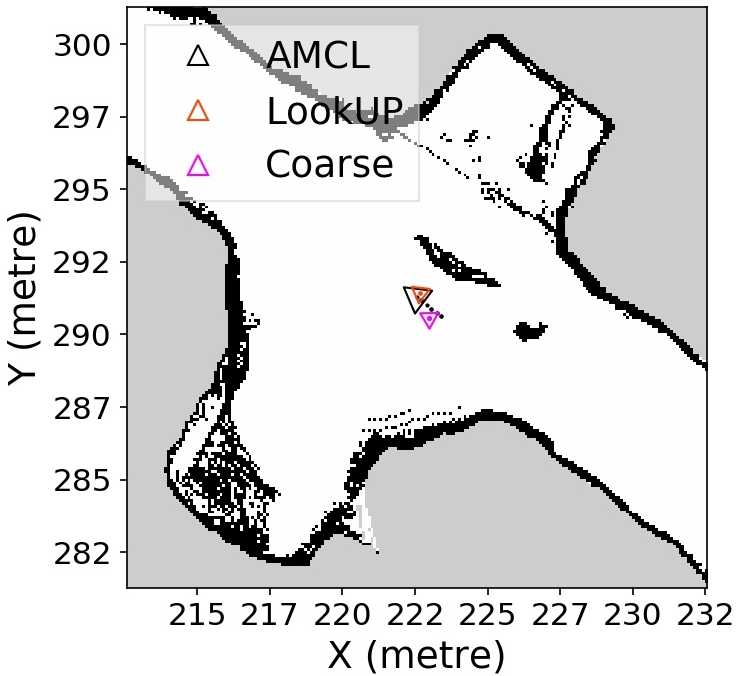}
		\caption{}
		\label{fig:localisation_result.canninton.pose_in_map_2}
	\end{subfigure}

	\caption{(a) Optical flow between the query image (top row) and various reference images (bottom row) for the frame in (d). (b-d) Localisation results of three sample frames from Mine B dataset, showing refinements in different directions.}
	\label{fig:localisation_result.cannington}
	\vspace{-10pt}
\end{figure}

As shown in  Fig.~\ref{fig:localisation_result.cannington.sample_frame}, \sysname{} can successfully extract optical flow in various directions and its ability to refine the coarse localisation results is not limited to the travel direction (Figs.~\ref{fig:localisation_result.canninton.pose_in_map_0}-\ref{fig:localisation_result.canninton.pose_in_map_2}).

\subsection{Effectiveness of Sample Point Classifier} \label{subsection:effectiveness_of_sample_point_classifier}

\begin{figure}[htp]
	\centering
	\includegraphics[width=0.4\textwidth]{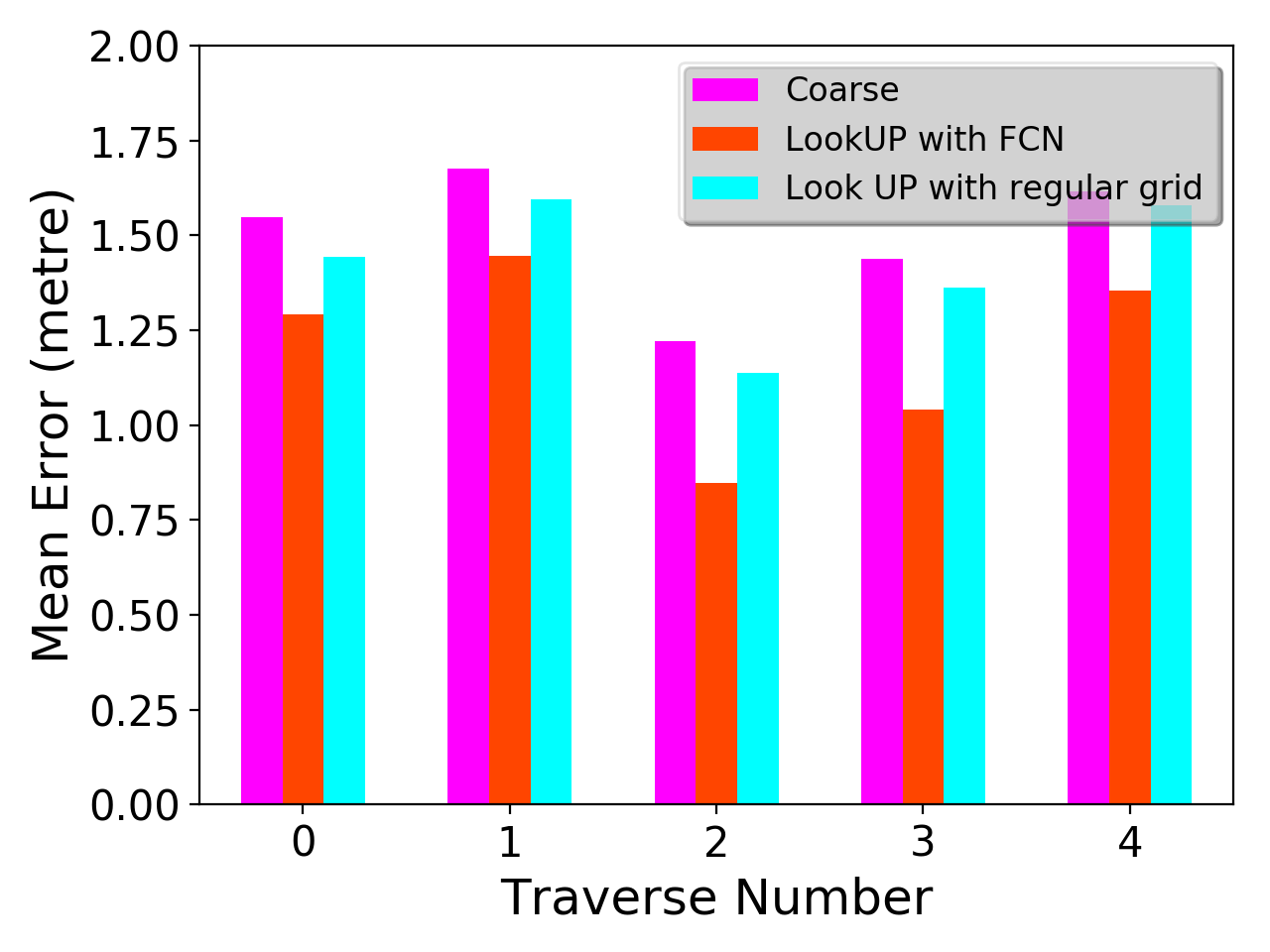}
	\caption{Mean localisation error for each query traverse under different system settings.}
	\label{fig:mean_error}
	\vspace{-10pt}
\end{figure}

The mean localisation errors obtained for each traverse with: a) \coarseLocalisationUnitName{} without refinement, b) \sysname{} with FCN and c) \sysname{} with regular grid sample point selector, are shown in Fig.~\ref{fig:mean_error}.
The localisation refinements computed by 
\sysname{} with regular grid leads to consistent but small error reductions, while \sysname{} with FCN sample point selector consistently leads to significant error reduction (as much as $\sim$27\% for traverse 3). 
This is because the indiscriminately sampled points on a regular grid resulted in false positive matches and therefore inaccurate optical flows for the Pixel Correspondence Matcher. Note that the mean errors reported for the coarse localisation method in  Fig.~\ref{fig:mean_error} are significantly lower than the $\sim$9.44 metres reported in~\cite{Jacobson2018}. There are two major reasons: Firstly, as mentioned previously, for all traverses and all methods, frames for which \coarseLocalisationUnitName{} produced errors greater than 10 metres were excluded from the evaluation. Secondly, the map and set of benchmarks used in~\cite{Jacobson2018} were generated by an external Radio Telemetry System, which was less accurate than AMCL.

\subsection{Computation Time}
To study computation time performance of the \sysname{} unit, the coarse localisation result was obtained first by running \coarseLocalisationUnitName{} unit with all the query images and the confusion matrix was saved before the timer was started. The following processes are all included in the computation time: for each query image, The \sysname{} reads the pre-computed confusion matrix, searches for the best-match coarse reference image from the forward-facing camera for that query, and ``looks up'' the corresponding ceiling images with the closest time stamp. The homography result is then calculated and saved as a file. Subsequent filtering, analyses and plotting are not timed. On an Intel i7-7700K 4.20GHz CPU, \sysname{} with FCN took 15 minutes to generate all results for Traverse 0 of Mine A, an averaged $\sim$\bestFrameRate{} frames per second (fps), which is acceptable for real-time operations in our application. Note there could be multiple reference images processed for each query input, the frame rate for processing each reference-query pair is $\sim$22 fps.
\section{Conclusion} \label{section:conclusion}

In this paper, we designed and characterised a refinement unit ``\sysname{}'' to our localisation system for vehicles in underground mine tunnel environments. It works by finding homographies based on matched pixels between query and reference images of the mine ceiling. The accuracy of \sysname{} is enhanced by generating pixel correspondences only on high-quality sample points proposed by an FCN. Selectively processing high-quality sample points also significantly increased the frame rate to $\sim$\bestFrameRate{} fps. This result was obtained using code that is yet to be optimised and could potentially be even faster if a GPU is available in the system.
The proposed system provides a viable framework for industrial applications in underground mines.







\bibliographystyle{IEEEtran}
\bibliography{publications}

\end{document}